\documentclass[11pt,a4paper]{article}
\usepackage[hyperref]{AACL-IJCNLP2020}
\usepackage{times}
\usepackage{latexsym}
\usepackage{graphicx}
\usepackage{caption}
\usepackage{subcaption}

\usepackage{amsfonts}
\usepackage{url}
\usepackage{xcolor}
\usepackage[ruled,linesnumbered]{algorithm2e}
\usepackage{graphicx}
\usepackage{bbm}

\usepackage{amsmath}

\usepackage{microtype}

\aclfinalcopy 

\title{Understanding Pre-trained BERT for Aspect-based Sentiment Analysis}

\author{Hu Xu\textsuperscript{\text{1}}, Lei Shu\textsuperscript{\text{1}}, Philip S. Yu\textsuperscript{\text{1,2}}\and Bing Liu\textsuperscript{\text{1}}\\
    \textsuperscript{1}{Department of Computer Science, University of Illinois at Chicago, Chicago, IL, USA}\\
    \textsuperscript{2}{Institute for Data Science, Tsinghua University, Beijing, China}\\
    {\tt \{hxu48,lshu3,psyu,liub\}@uic.edu}
}

\date{}

\begin{document}
\maketitle
\begin{abstract}
This paper analyzes the pre-trained hidden representations learned from reviews on BERT for tasks in aspect-based sentiment analysis (ABSA).
Our work is motivated by the recent progress in BERT-based language models for ABSA.
However, it is not clear how the general proxy task of (masked) language model trained on unlabeled corpus without annotations of aspects or opinions can provide important features for downstream tasks in ABSA. 
By leveraging the annotated datasets in ABSA, we investigate both the attentions and the learned representations of BERT pre-trained on reviews. 
We found that BERT uses very few self-attention heads to encode context words (such as prepositions or pronouns that indicating an aspect) and opinion words for an aspect.
Most features in the representation of an aspect are dedicated to the fine-grained semantics of the domain (or product category) and the aspect itself, instead of carrying summarized opinions from its context.
~We hope this investigation can help future research in improving self-supervised learning, unsupervised learning and fine-tuning for ABSA.
\footnote{The pre-trained model and code can be found at \url{https://github.com/howardhsu/BERT-for-RRC-ABSA}.}
\end{abstract}

\section{Introduction}
As a form of self-supervised learning in NLP, pre-trained language models (LMs) like the masked LM in BERT \cite{devlin2018bert,liu2019roberta,lan2019albert} yield significant performance gains when later fine-tuned on downstream NLP tasks.
Recent studies also showed impressive results on tasks in aspect-based sentiment analysis (ABSA) \cite{xu_bert2019,sun-etal-2019-utilizing,li2019exploiting,tian-etal-2020-skep,karimi2020adversarial}, which aims to discover aspects and their associated opinions \cite{hu2004mining,liu2012sentiment}.
Although there are existing studies of the hidden representations and attentions of LMs about tasks such as parsing and co-reference resolution \cite{adi2016finegrained,belinkov-etal-2017-neural,clark2019does}, 
it is unclear how LMs capture aspects and sentiment/opinion from large-scale unlabeled texts.

This paper attempts to investigate and understand the inner workings of the pretext task of the masked language model (MLM) in transformer and their connections with tasks in ABSA.
This may benefit the following problems: (1) improving fine-tuning of ABSA if we have a better understanding about the gap between pretext tasks and fine-tuning tasks; (2) more importantly, self-supervised (or unsupervised) ABSA without fine-tuning to save the expensive efforts on annotating ABSA datasets (for a new domain).

We are particularly interested in fine-grained token-level features that are typically required by ABSA and how MLM as a general task can cover them during pre-training.
Typical tasks of ABSA are: aspect extraction (AE), aspect sentiment classification (ASC) \cite{hu2004mining,dong2014adaptive,nguyen-shirai-2015-phrasernn,li2018transformation,tang2016aspect,wang2016recursive,wang2016attention,ma2017interactive,chen2017recurrent,ma2017interactive,tay2018learning,he2018effective,liu2018content} and end-to-end ABSA (E2E-ABSA) \cite{li2019unified,li2019exploiting}.
AE aims to extract aspects (e.g., ``battery'' in the \textit{laptop} domain), ASC identifies the polarity for a given aspect (e.g., \emph{positive} about \emph{battery}) and E2E-ABSA is a combination of AE and ASC that detects aspects and their associated polarities simultaneously.
Existing studies show that tasks in ABSA require the understanding of the interactions of aspects (e.g., ``screen'' in the \textit{laptop} domain) and its contexts, including sentiment (e.g., ``clear'') \cite{qiu2011opinion,wang2017coupled,he_acl2019}.
As such, we believe how the hidden representation of an aspect encodes features about being an aspect and summarizing opinions of that aspect are crucial for ABSA.


This paper represents a new addition to the existing analysis of BERT in~\cite{clark2019does}, which focuses on studying the behavior of BERT's hidden representation and self-attentions for general purposes.
We focus on how the self-supervised training of BERT is prepared for fine-grained features that are important for ABSA. 
We leverage the annotated data of ABSA in our analysis to draw the relevance between pre-trained features and labels for ABSA.
Note that we do not use any of such annotations for fine-tuning or training.
Also, we do not study LMs that carry extra human supervision for sentiment analysis (such as using opinion lexicons as in \cite{ke2019sentilr,tian-etal-2020-skep}) because we are interested in to what degree a general self-supervised task such as MLM can cover specific tasks in ABSA.
Unlike \cite{clark2019does,sun2019utilizing,wan2020target}, we focus on masked language model (MLM) for fine-grained token-level features instead of next sentence prediction (NSP) since the latter is not widely adopted for pre-trained models~\cite{liu2019roberta}.

Our main finding is that BERT (pre-trained on reviews) encodes rich semantic knowledge about the domain and aspect itself into the hidden representations of an aspect but uses almost no dedicated features for opinions.
Inside BERT, 
very few self-attention heads are learned for encoding salient context words for finding an aspect or summarizing opinion words for an aspect.
This suggests the pros and cons of MLM.
For example, predicting a masked aspect word to learn features for an aspect is a weak self-supervised learning task for ABSA.
This leads to future directions on alternative self-supervised learning tasks for ABSA.

\section{Pre-trained LMs and Datasets}
We expect to simplify our analysis on the same latent space for multiple domains of ABSA.
As such, we pre-train BERT on reviews with large coverage of domains (product categories).
The training corpus is a combination of Amazon reviews \cite{HeMcA16a} and Yelp review datasets \footnote{\url{https://www.yelp.com/dataset/challenge}}, which give us a review corpus of 20+ GB in size.
We start from fine-tuning $\textsc{BERT}_{\text{base}}$ on such a corpus for 4 epochs.
We train it on a single TITAN RTX GPU for 10 days. 

To understand the hidden representation of an aspect and its formulation from self-attention, we leverage the popular SemEval 2014 Task 4 and SemEval-2016 Task 5 in ABSA with annotations about aspects and their associated opinions.
These benchmark datasets cover the domains of \textit{Laptop} and  \textit{Restaurant}. We sample 150 examples from each domain as the validation data for analysis. Note that we do not use any annotated data for fine-tuning or training because we are interested in how relevant the features from pre-trained BERT to ABSA.

\section{Analysis}

To see the inner workings of masked language modeling (MLM) on reviews, 
we first review MLM and transformers. Then we perform two types of evaluations: self-attention of aspects and hidden representations on aspects.

Transformer \cite{vaswani2017attention} is a neural architecture that purely uses multi-head self-attentions to learn hidden representations of input texts.
Unlike the classic LSTM or CNN, the connections in self-attention can be viewed as a fully-connected graph on nodes of tokens without strong inductive bias on contexts or prior knowledge on relative positions. So transformer uses positional embeddings to encode tokens at different positions. Further, multi-head attentions can be viewed as typed relations to model various kinds of relations among tokens. 

MLMs aim to recover texts corrupted with masked tokens. For example, from ``The \texttt{[MASK]} is clear.'' in the laptop domain, one can easily guess that \texttt{[MASK]} is probably ``screen''. As a result, the transformer model must use self-attentions to infer the hidden representation of ``screen'' from ``The'', ``is'' and ``clear''. 
Note that the first embedding layer of BERT and the MLM prediction heads are both context-independent because they just contain word embeddings, whereas the other internal layers in-between are context-dependent.
As such, how BERT encodes such contexts into the representation of an aspect is important for ABSA. 

\subsection{Self-Attentions of Aspects}

\begin{figure*}[!t]
    \begin{subfigure}[b]{0.15\textwidth}
    \includegraphics[width=\textwidth]{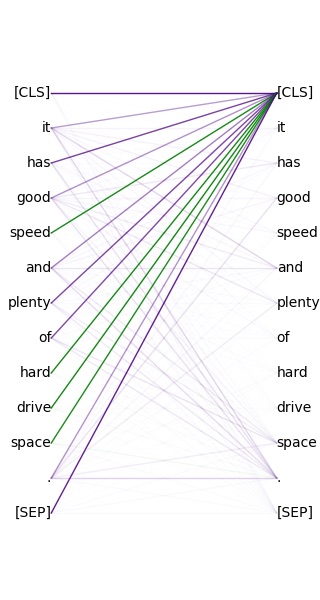}
    \caption{no-op}
    \end{subfigure}
    \hfill
    \begin{subfigure}[b]{0.15\textwidth}
    \includegraphics[width=\textwidth]{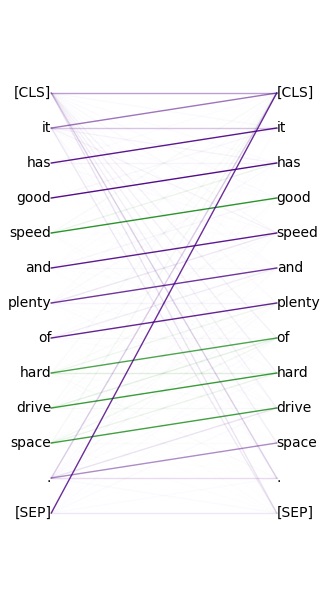}
    \caption{offset}
    \end{subfigure}
    \hfill
    \begin{subfigure}[b]{0.15\textwidth}
    \includegraphics[width=\textwidth]{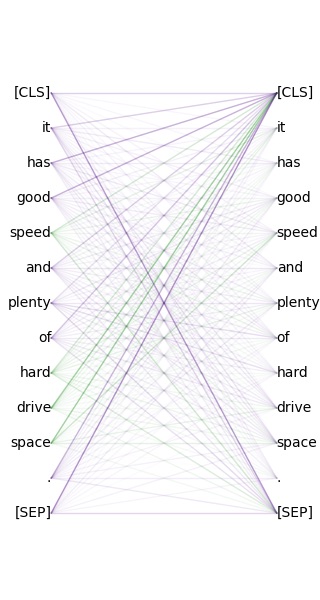}
    \caption{broadcast}
    \end{subfigure}
    \hfill
    \begin{subfigure}[b]{0.15\textwidth}
    \includegraphics[width=\textwidth]{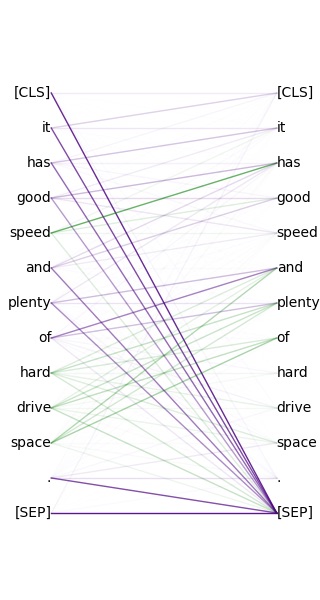}
    \caption{context}
    \end{subfigure}
    \hfill
    \hfill
    \begin{subfigure}[b]{0.15\textwidth}
    \includegraphics[width=\textwidth]{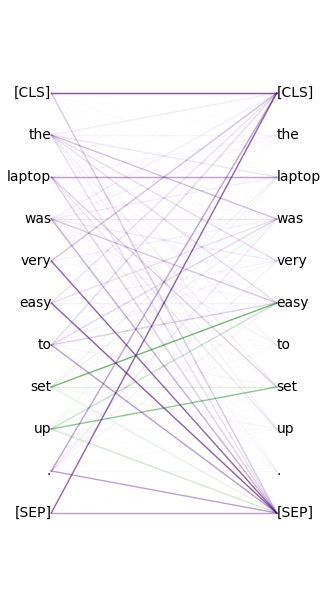}
    \caption{opinion}
    \end{subfigure}
    \hfill
    \begin{subfigure}[b]{0.15\textwidth}
    \includegraphics[width=\textwidth]{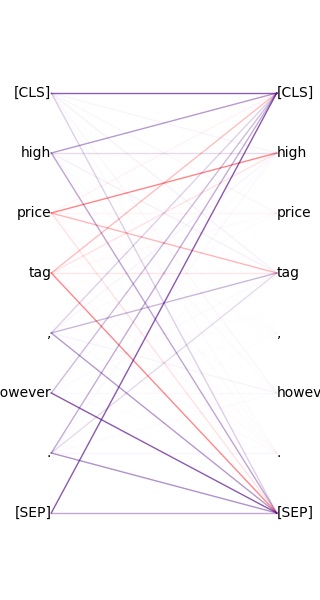}
    \caption{opinion}
    \end{subfigure}
\caption{Self-attentions (the darkness of a line indicates the strength of attention): (a) no-op attentions on the \texttt{[CLS]} or \texttt{[SEP]} token; (b) offset attentions to encode local contexts; (c) broadcasting attentions to encode global contexts; (d) context words of an aspect (e.g., ``has'') and (e-f) opinion words: green and red lines indicate positive and negative opinions, respectively (best viewed in colors).}
\label{fig:sa}
\vspace{-3mm}
\end{figure*}

\begin{figure*}[t]
\centering
    \begin{subfigure}[b]{3.1in}
    \includegraphics[width=3.1in]{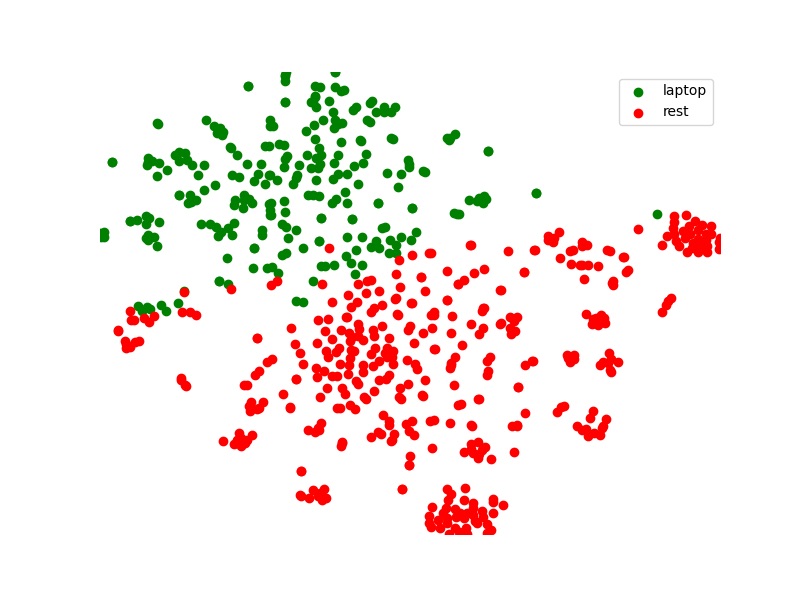}
    \caption{\textit{laptop(green)} vs \textit{restaurant(red)}}
    \end{subfigure}
    \hfill
    \begin{subfigure}[b]{3.1in}
    \includegraphics[width=3.1in]{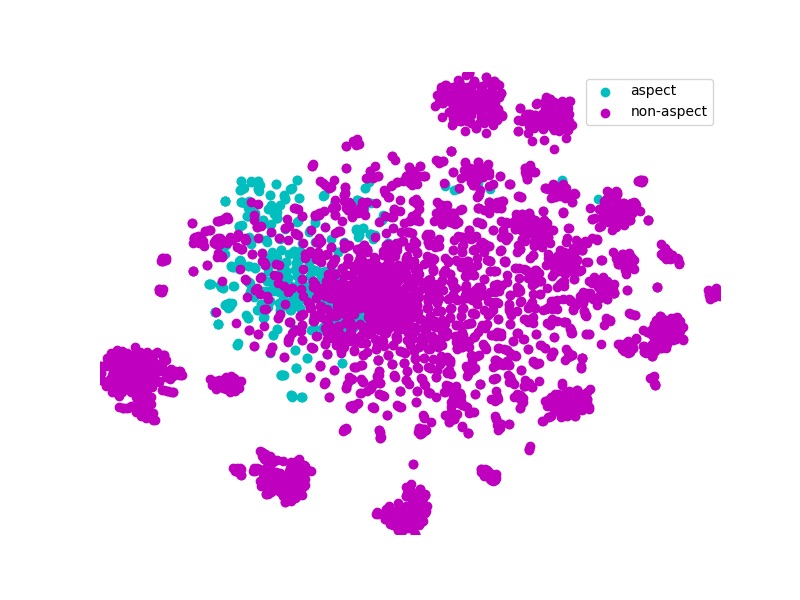}
    \caption{aspects (blue) vs non-aspects (purple)}
    \end{subfigure}
    \caption{t-SNE of hidden representations of aspects on domains and aspect vs non-aspect words (best viewed in colors).}
\label{fig:ana}
\vspace{-3mm}
\end{figure*}

Intuitively, self-attention can serve as a way to aggregate the representations of contextual tokens into an aspect.
Following \cite{clark2019does}, we notice some attention heads exhibit general patterns such as (a) no-op on \texttt{[CLS]} or \texttt{[SEP]}, (b) offsets on previous/next tokens and (c) broadcasting over the whole sentence, as in Figure \ref{fig:sa}.
For pattern (a), transformer has to take no-ops on redundant tokens given that the softmax function in self-attention always normalized to 1 even when no relevant context tokens are presented.
For pattern (b), self-attention needs recurrent patterns to construct local contexts from nearby tokens.
Pattern (c) can be viewed as a global average pooling operation so that each token contains knowledge from the whole sequence.

We are also interested in how an aspect interacts with its contexts, including both context words that indicate an aspect (e.g., ``of'' and ``has'') and opinion words (e.g., ``good'').
We search through all 144 heads in $\textsc{BERT}_{base}$ but only find 2 to 4 heads of such types residing in the middle layers.
These self-attentions could help infer a \texttt{[MASK]}ed aspect in MLM. 
For example, in ``The \texttt{[MASK]} is clear'', the representation of ``clear'' may contain important features and later be aggregated to the representation of \texttt[MASK] to predict the word ``screen''. 
However, BERT may need to erase the semantic of \textit{positive} in ``clear'' when formulating the representation of \texttt{[MASK]}.
This is because the last layer (prediction head of MLM) are context-independent and the word embedding of ``screen'' is expected to have no sentiment.
This may hurt the representations of aspect words to carry sentiment. 

\subsection{Hidden Representations on Aspects}
\begin{figure*}[!t]
\centering    
    \begin{subfigure}[b]{3.1in}
    \centering
    \includegraphics[width=3.1in]{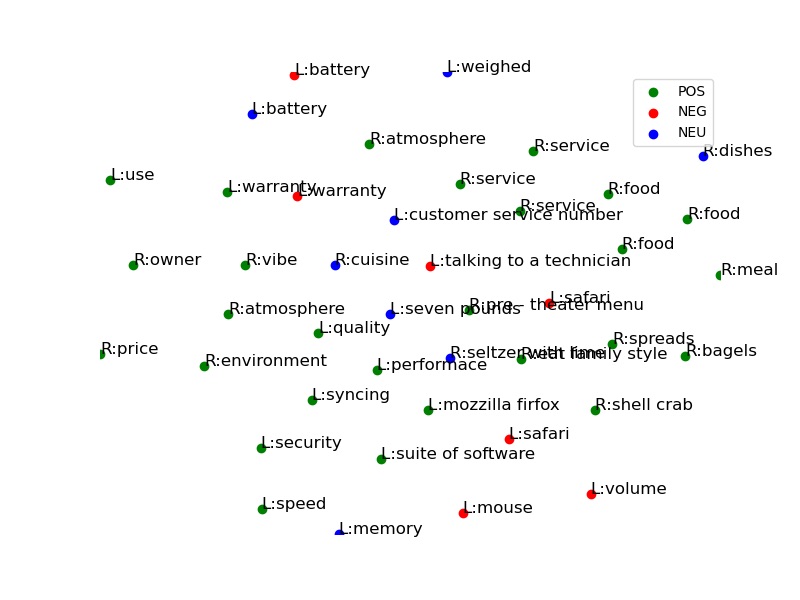}
    \caption{\textit{laptop}(L) and \textit{restaurant}(R), with their associated opinions}
    \vspace{-3mm}
    \end{subfigure}
    \hfill
    \begin{subfigure}[b]{3.1in}
    \centering
    \includegraphics[width=3.1in]{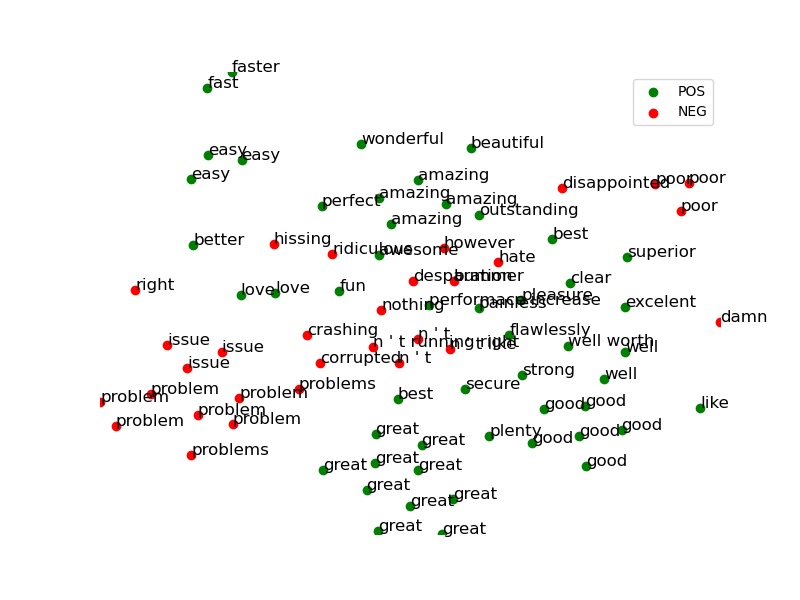}
    \caption{opinion words}
    \vspace{-3mm}
    \end{subfigure}
    \caption{t-SNE of hidden representations of aspects and opinion words (best viewed in colors).}
\label{fig:aspect}
\vspace{-5mm}
\end{figure*}

\begin{figure*}[!t]
\centering    
\begin{subfigure}[b]{3.1in}
    \centering
    \includegraphics[width=3.4in]{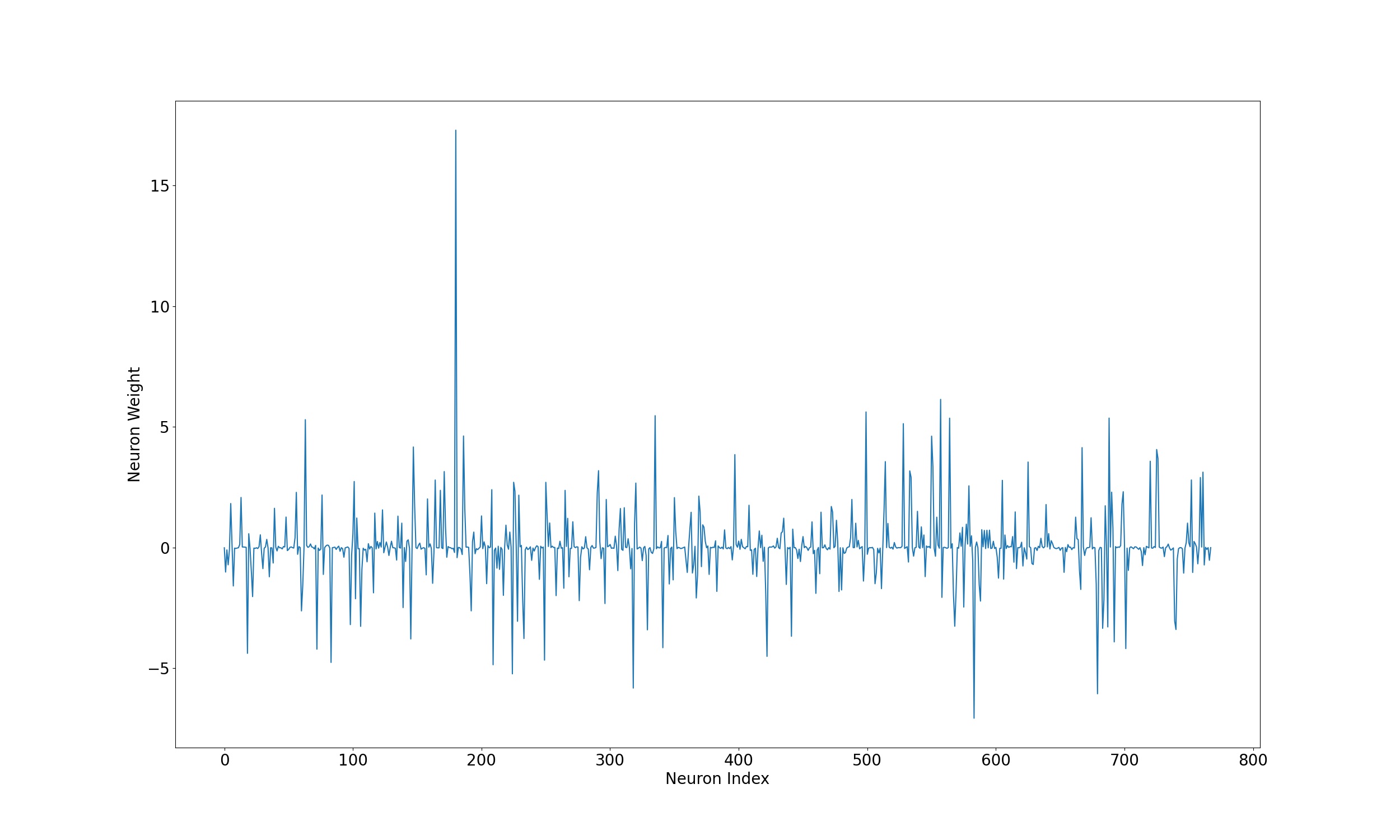}
    \caption{aspect and non-aspect on both \textit{laptop}.}
    \vspace{-3mm}
    \end{subfigure}
    \hfill
    \begin{subfigure}[b]{3.1in}
    \centering
    \includegraphics[width=3.4in]{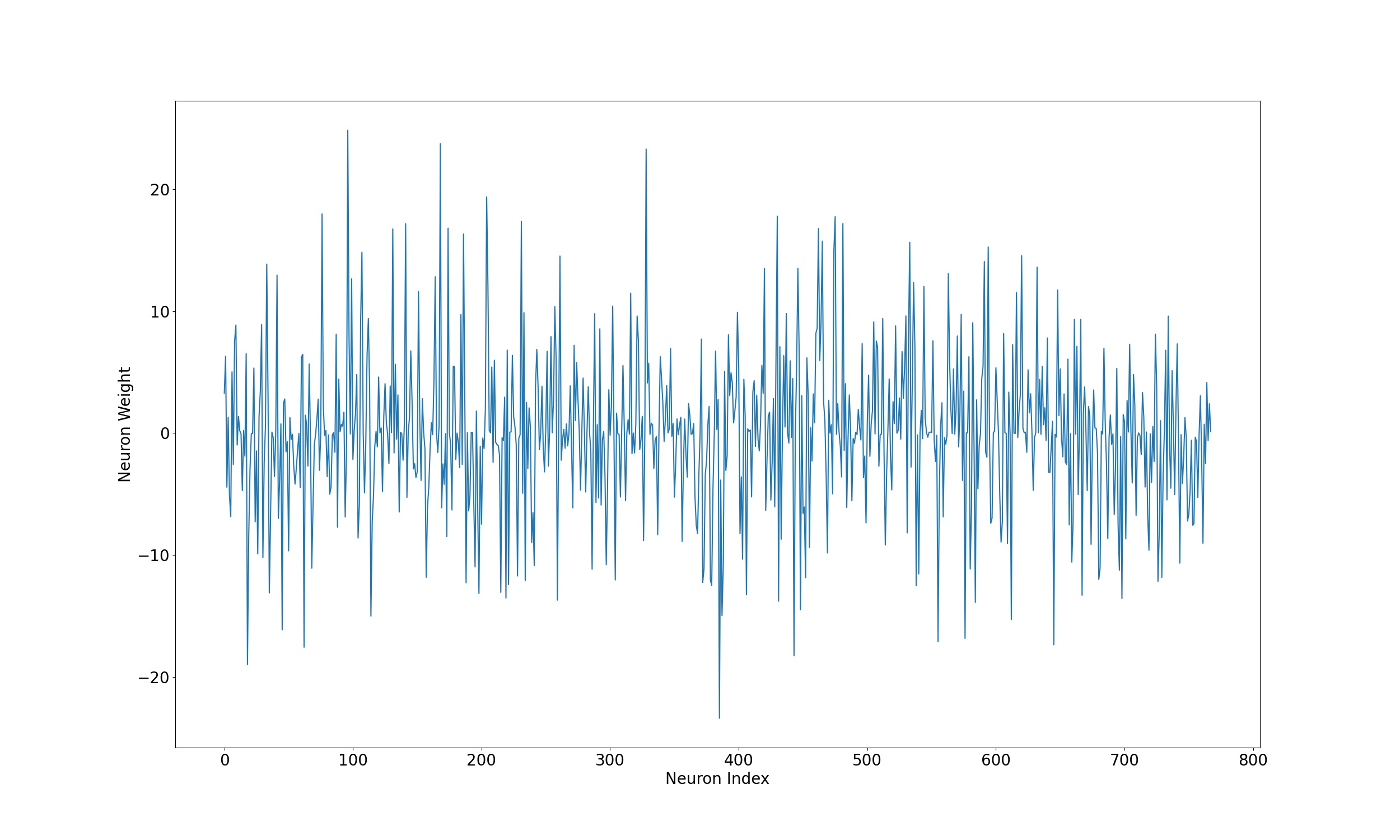}
    \caption{sentiment polarity}
    \vspace{-3mm}
    \end{subfigure}
    \caption{L1-normed probe on 768 neurons (size of hidden space) of BERT: neuron 180 stands out on (a) but not salient enough.}
    \vspace{-3mm}
\label{fig:probe}
\end{figure*}

Next, we analyze aspects in the latent space.
First, since tasks in ABSA are strongly domain-dependent, we are interested in how much domain knowledge is carried in aspects.
With aspects under t-SNE \cite{maaten2008visualizing} dimensionality reduction as in Figure \ref{fig:ana}(a), we notice that BERT exhibits a strong separation of domains in its latent space of aspects. This indicates that BERT devotes most of its feature dimensions to domain differences.
We plot the hidden representations of the last layer for tokens about aspects and non-aspects in Figure \ref{fig:ana}(b).
We can see that in the 2D space of t-SNE, aspects and non-aspect tokens occupy slightly different spaces, indicating a great number of dimensions in the latent space are devoted to features that can be used to separate aspects from its contexts.
This correlates well with existing research showing that BERT has impressive results on aspect extraction \cite{xu_bert2019}.

We further investigate whether there exists a single general (cross-domain) neuron to separate aspects from other words.
This is important for unsupervised or zero-shot learning for aspect extraction because if one can easily adapt a pre-trained BERT for a new domain without annotated fine-tuning data.
We train an L1-normed probe on hidden states of aspects and non-aspect words from both domains, as depicted in Figure \ref{fig:probe}(a).
The neuron 180 stands out from other neurons.
To test whether this neuron is salient or general enough,
we use this neuron alone as the single feature for aspect word classification (extraction).
The F1 score is 31\%, whereas using all neurons reaches 79\%.
This implies no salient neuron is available for general-purpose aspect extraction.

Further, we are interested in the relations between aspects and their associated opinions.
Since an aspect may contain multiple words (``customer service'') and a word may be tokenized into multiple tokens in BERT (``warrant'' and ``\#\#y''), we average the representations of tokens belonging to one aspect.
From Figure \ref{fig:aspect}, we can see that the dimensions that dominate the hidden space of aspects are their semantic meanings, where aspects with similar meanings are closer. 
For example, different kinds of software on laptop are closer. 
The differences in opinion do not exhibit significant impacts on the representation of an aspect, indicating few dimensions are dedicated to opinions. 

To verify the existence of features contributing to opinions, we train a (logistic regression) probe to classify sentiment (positive or negative) over the frozen latent space of aspects, with L1-norm on weights to pick salient features.
We obtain an F1-score of 83\% on the polarities of aspects in the test set. In Figure \ref{fig:probe}(b),
one can observe that no single neuron stands out as a feature for sentiment classification on aspects but many features are correlated with the sentiment. 
This is in contrast with the findings of sentiment neuron in pre-trained casual LMs \cite{radford2017learning} and we suspect the small hidden space (768) does not allow such a single neuron to exist.
This indicates that MLM is a weak task for summarizing and carrying sentiment polarities to aspects (e.g., E2E-ABSA).

We further explore the hidden representations of opinion words (or opinion terms (OT)), as in Figure \ref{fig:aspect}(b).
One can observe that opinion words of different polarities do not have their very own sub-spaces but occupy similar spaces in general. 
Instead, they carry their non-opinion meanings (such as the semantic of speed in ``fast'') and reflect such differences in the latent space.

\section{Summary and Open Problems}
Our analysis showed that MLM tends to learn very fine-grained features and dedicate most of the aspects' features to domains and semantics of aspects themselves rather than opinions.
We believe using pre-trained BERT is good for AE or the extraction part of E2E-ABSA but poor for summarizing opinions in ASC or polarity detection in E2E-ABSA.
End tasks may still require a good amount of examples to explore the large feature space of BERT.

We believe this analysis leads to the following open problems: alternative self-supervised learning tasks besides MLM. We believe MLM is far from perfect as a pretext task for ABSA and both learned features for AE and ASC can be improved. 
We believe the main weakness of MLM is that when learning representations for an aspect word, it does not need to know the sentiment in most cases and being an aspect or not is not a strong feature to learn MLM.
How to design a pretext task to learn disentangle aspect features and opinion features from fine-grained features of other semantics is still an open problem. 
For aspect features, one may group reviews for the same item (e.g., a product) to encourage the signal for aspects against other items.
For sentiment features, one may use rating as a weakly supervised signal to strengthen aspect words with sentiment.

\section*{Acknowledgments}
This work was partially supported by four grants from National Science Foundation: IIS-1910424, IIS-1838770, III-1763325, and III-1909323, a DARPA Contract HR001120C0023, and SaTC-1930941.

\bibliography{anthology,aacl-ijcnlp2020}

\begin{thebibliography}{35}
\expandafter\ifx\csname natexlab\endcsname\relax\def\natexlab#1{#1}\fi

\bibitem[{Adi et~al.(2016)Adi, Kermany, Belinkov, Lavi, and
  Goldberg}]{adi2016finegrained}
Yossi Adi, Einat Kermany, Yonatan Belinkov, Ofer Lavi, and Yoav Goldberg. 2016.
\newblock \href {http://arxiv.org/abs/1608.04207} {Fine-grained analysis of
  sentence embeddings using auxiliary prediction tasks}.

\bibitem[{Belinkov et~al.(2017)Belinkov, Durrani, Dalvi, Sajjad, and
  Glass}]{belinkov-etal-2017-neural}
Yonatan Belinkov, Nadir Durrani, Fahim Dalvi, Hassan Sajjad, and James Glass.
  2017.
\newblock \href {https://doi.org/10.18653/v1/P17-1080} {What do neural machine
  translation models learn about morphology?}
\newblock In \emph{Proceedings of the 55th Annual Meeting of the Association
  for Computational Linguistics (Volume 1: Long Papers)}, pages 861--872,
  Vancouver, Canada. Association for Computational Linguistics.

\bibitem[{Chen et~al.(2017)Chen, Sun, Bing, and Yang}]{chen2017recurrent}
Peng Chen, Zhongqian Sun, Lidong Bing, and Wei Yang. 2017.
\newblock Recurrent attention network on memory for aspect sentiment analysis.
\newblock In \emph{Proceedings of the 2017 conference on empirical methods in
  natural language processing}, pages 452--461.

\bibitem[{Clark et~al.(2019)Clark, Khandelwal, Levy, and
  Manning}]{clark2019does}
Kevin Clark, Urvashi Khandelwal, Omer Levy, and Christopher~D Manning. 2019.
\newblock What does bert look at? an analysis of bert's attention.
\newblock \emph{arXiv preprint arXiv:1906.04341}.

\bibitem[{Devlin et~al.(2019)Devlin, Chang, Lee, and
  Toutanova}]{devlin2018bert}
Jacob Devlin, Ming-Wei Chang, Kenton Lee, and Kristina Toutanova. 2019.
\newblock \href {https://doi.org/10.18653/v1/N19-1423} {{BERT}: Pre-training of
  deep bidirectional transformers for language understanding}.
\newblock In \emph{Proceedings of the 2019 Conference of the North {A}merican
  Chapter of the Association for Computational Linguistics: Human Language
  Technologies, Volume 1 (Long and Short Papers)}, pages 4171--4186,
  Minneapolis, Minnesota. Association for Computational Linguistics.

\bibitem[{Dong et~al.(2014)Dong, Wei, Tan, Tang, Zhou, and
  Xu}]{dong2014adaptive}
Li~Dong, Furu Wei, Chuanqi Tan, Duyu Tang, Ming Zhou, and Ke~Xu. 2014.
\newblock Adaptive recursive neural network for target-dependent twitter
  sentiment classification.
\newblock In \emph{Proceedings of the 52nd annual meeting of the association
  for computational linguistics (volume 2: Short papers)}, volume~2, pages
  49--54.

\bibitem[{He et~al.()He, Lee, Ng, and Dahlmeier}]{he_acl2019}
Ruidan He, Wee~Sun Lee, Hwee~Tou Ng, and Daniel Dahlmeier.
\newblock An interactive multi-task learning network for end-to-end
  aspect-based sentiment analysis.
\newblock In \emph{Proceedings of the 57th Annual Meeting of the Association
  for Computational Linguistics}. Association for Computational Linguistics.

\bibitem[{He et~al.(2018)He, Lee, Ng, and Dahlmeier}]{he2018effective}
Ruidan He, Wee~Sun Lee, Hwee~Tou Ng, and Daniel Dahlmeier. 2018.
\newblock Effective attention modeling for aspect-level sentiment
  classification.
\newblock In \emph{Proceedings of the 27th International Conference on
  Computational Linguistics}, pages 1121--1131.

\bibitem[{He and McAuley(2016)}]{HeMcA16a}
Ruining He and Julian McAuley. 2016.
\newblock Ups and downs: Modeling the visual evolution of fashion trends with
  one-class collaborative filtering.
\newblock In \emph{World Wide Web}.

\bibitem[{Hu and Liu(2004)}]{hu2004mining}
Minqing Hu and Bing Liu. 2004.
\newblock Mining and summarizing customer reviews.
\newblock In \emph{Proceedings of the tenth ACM SIGKDD international conference
  on Knowledge discovery and data mining}, pages 168--177. ACM.

\bibitem[{Karimi et~al.(2020)Karimi, Rossi, Prati, and
  Full}]{karimi2020adversarial}
Akbar Karimi, Leonardo Rossi, Andrea Prati, and Katharina Full. 2020.
\newblock Adversarial training for aspect-based sentiment analysis with bert.
\newblock \emph{arXiv preprint arXiv:2001.11316}.

\bibitem[{Ke et~al.(2019)Ke, Ji, Liu, Zhu, and Huang}]{ke2019sentilr}
Pei Ke, Haozhe Ji, Siyang Liu, Xiaoyan Zhu, and Minlie Huang. 2019.
\newblock Sentilr: Linguistic knowledge enhanced language representation for
  sentiment analysis.
\newblock \emph{arXiv preprint arXiv:1911.02493}.

\bibitem[{Lan et~al.(2019)Lan, Chen, Goodman, Gimpel, Sharma, and
  Soricut}]{lan2019albert}
Zhenzhong Lan, Mingda Chen, Sebastian Goodman, Kevin Gimpel, Piyush Sharma, and
  Radu Soricut. 2019.
\newblock Albert: A lite bert for self-supervised learning of language
  representations.
\newblock In \emph{International Conference on Learning Representations}.

\bibitem[{Li et~al.(2018)Li, Bing, Lam, and Shi}]{li2018transformation}
Xin Li, Lidong Bing, Wai Lam, and Bei Shi. 2018.
\newblock Transformation networks for target-oriented sentiment classification.
\newblock \emph{arXiv preprint arXiv:1805.01086}.

\bibitem[{Li et~al.(2019{\natexlab{a}})Li, Bing, Li, and Lam}]{li2019unified}
Xin Li, Lidong Bing, Piji Li, and Wai Lam. 2019{\natexlab{a}}.
\newblock A unified model for opinion target extraction and target sentiment
  prediction.
\newblock In \emph{Proceedings of the AAAI Conference on Artificial
  Intelligence}, volume~33, pages 6714--6721.

\bibitem[{Li et~al.(2019{\natexlab{b}})Li, Bing, Zhang, and
  Lam}]{li2019exploiting}
Xin Li, Lidong Bing, Wenxuan Zhang, and Wai Lam. 2019{\natexlab{b}}.
\newblock Exploiting bert for end-to-end aspect-based sentiment analysis.
\newblock \emph{arXiv preprint arXiv:1910.00883}.

\bibitem[{Liu(2012)}]{liu2012sentiment}
Bing Liu. 2012.
\newblock Sentiment analysis and opinion mining.
\newblock \emph{Synthesis lectures on human language technologies},
  5(1):1--167.

\bibitem[{Liu et~al.(2018)Liu, Zhang, Zeng, Huang, and Wu}]{liu2018content}
Qiao Liu, Haibin Zhang, Yifu Zeng, Ziqi Huang, and Zufeng Wu. 2018.
\newblock Content attention model for aspect based sentiment analysis.
\newblock In \emph{Proceedings of the 2018 World Wide Web Conference on World
  Wide Web}, pages 1023--1032. International World Wide Web Conferences
  Steering Committee.

\bibitem[{Liu et~al.(2019)Liu, Ott, Goyal, Du, Joshi, Chen, Levy, Lewis,
  Zettlemoyer, and Stoyanov}]{liu2019roberta}
Yinhan Liu, Myle Ott, Naman Goyal, Jingfei Du, Mandar Joshi, Danqi Chen, Omer
  Levy, Mike Lewis, Luke Zettlemoyer, and Veselin Stoyanov. 2019.
\newblock Roberta: A robustly optimized bert pretraining approach.
\newblock \emph{arXiv preprint arXiv:1907.11692}.

\bibitem[{Ma et~al.(2017)Ma, Li, Zhang, and Wang}]{ma2017interactive}
Dehong Ma, Sujian Li, Xiaodong Zhang, and Houfeng Wang. 2017.
\newblock Interactive attention networks for aspect-level sentiment
  classification.
\newblock \emph{arXiv preprint arXiv:1709.00893}.

\bibitem[{Maaten and Hinton(2008)}]{maaten2008visualizing}
Laurens van~der Maaten and Geoffrey Hinton. 2008.
\newblock Visualizing data using t-sne.
\newblock \emph{Journal of machine learning research}, 9(Nov):2579--2605.

\bibitem[{Nguyen and Shirai(2015)}]{nguyen-shirai-2015-phrasernn}
Thien~Hai Nguyen and Kiyoaki Shirai. 2015.
\newblock \href {https://doi.org/10.18653/v1/D15-1298} {{P}hrase{RNN}: Phrase
  recursive neural network for aspect-based sentiment analysis}.
\newblock In \emph{Proceedings of the 2015 Conference on Empirical Methods in
  Natural Language Processing}, pages 2509--2514, Lisbon, Portugal. Association
  for Computational Linguistics.

\bibitem[{Qiu et~al.(2011)Qiu, Liu, Bu, and Chen}]{qiu2011opinion}
Guang Qiu, Bing Liu, Jiajun Bu, and Chun Chen. 2011.
\newblock Opinion word expansion and target extraction through double
  propagation.
\newblock \emph{Computational linguistics}, 37(1):9--27.

\bibitem[{Radford et~al.(2017)Radford, Jozefowicz, and
  Sutskever}]{radford2017learning}
Alec Radford, Rafal Jozefowicz, and Ilya Sutskever. 2017.
\newblock Learning to generate reviews and discovering sentiment.
\newblock \emph{arXiv preprint arXiv:1704.01444}.

\bibitem[{Sun et~al.(2019{\natexlab{a}})Sun, Huang, and
  Qiu}]{sun-etal-2019-utilizing}
Chi Sun, Luyao Huang, and Xipeng Qiu. 2019{\natexlab{a}}.
\newblock \href {https://doi.org/10.18653/v1/N19-1035} {Utilizing {BERT} for
  aspect-based sentiment analysis via constructing auxiliary sentence}.
\newblock In \emph{Proceedings of the 2019 Conference of the North {A}merican
  Chapter of the Association for Computational Linguistics: Human Language
  Technologies, Volume 1 (Long and Short Papers)}, pages 380--385, Minneapolis,
  Minnesota. Association for Computational Linguistics.

\bibitem[{Sun et~al.(2019{\natexlab{b}})Sun, Huang, and Qiu}]{sun2019utilizing}
Chi Sun, Luyao Huang, and Xipeng Qiu. 2019{\natexlab{b}}.
\newblock Utilizing bert for aspect-based sentiment analysis via constructing
  auxiliary sentence.
\newblock In \emph{Proceedings of the 2019 Conference of the North American
  Chapter of the Association for Computational Linguistics: Human Language
  Technologies, Volume 1 (Long and Short Papers)}, pages 380--385.

\bibitem[{Tang et~al.(2016)Tang, Qin, and Liu}]{tang2016aspect}
Duyu Tang, Bing Qin, and Ting Liu. 2016.
\newblock Aspect level sentiment classification with deep memory network.
\newblock \emph{arXiv preprint arXiv:1605.08900}.

\bibitem[{Tay et~al.(2018)Tay, Tuan, and Hui}]{tay2018learning}
Yi~Tay, Luu~Anh Tuan, and Siu~Cheung Hui. 2018.
\newblock Learning to attend via word-aspect associative fusion for
  aspect-based sentiment analysis.
\newblock In \emph{Thirty-Second AAAI Conference on Artificial Intelligence}.

\bibitem[{Tian et~al.(2020)Tian, Gao, Xiao, Liu, He, Wu, Wang, and
  wu}]{tian-etal-2020-skep}
Hao Tian, Can Gao, Xinyan Xiao, Hao Liu, Bolei He, Hua Wu, Haifeng Wang, and
  feng wu. 2020.
\newblock \href {https://www.aclweb.org/anthology/2020.acl-main.374} {{SKEP}:
  Sentiment knowledge enhanced pre-training for sentiment analysis}.
\newblock In \emph{Proceedings of the 58th Annual Meeting of the Association
  for Computational Linguistics}, pages 4067--4076, Online. Association for
  Computational Linguistics.

\bibitem[{Vaswani et~al.(2017)Vaswani, Shazeer, Parmar, Uszkoreit, Jones,
  Gomez, Kaiser, and Polosukhin}]{vaswani2017attention}
Ashish Vaswani, Noam Shazeer, Niki Parmar, Jakob Uszkoreit, Llion Jones,
  Aidan~N Gomez, {\L}ukasz Kaiser, and Illia Polosukhin. 2017.
\newblock Attention is all you need.
\newblock In \emph{Advances in neural information processing systems}, pages
  5998--6008.

\bibitem[{Wan et~al.(2020)Wan, Yang, Du, Liu, Qi, and Pan}]{wan2020target}
Hai Wan, Yufei Yang, Jianfeng Du, Yanan Liu, Kunxun Qi, and Jeff~Z Pan. 2020.
\newblock Target-aspect-sentiment joint detection for aspect-based sentiment
  analysis.
\newblock In \emph{AAAI}, pages 9122--9129.

\bibitem[{Wang et~al.(2016{\natexlab{a}})Wang, Pan, Dahlmeier, and
  Xiao}]{wang2016recursive}
Wenya Wang, Sinno~Jialin Pan, Daniel Dahlmeier, and Xiaokui Xiao.
  2016{\natexlab{a}}.
\newblock Recursive neural conditional random fields for aspect-based sentiment
  analysis.
\newblock \emph{arXiv preprint arXiv:1603.06679}.

\bibitem[{Wang et~al.(2017)Wang, Pan, Dahlmeier, and Xiao}]{wang2017coupled}
Wenya Wang, Sinno~Jialin Pan, Daniel Dahlmeier, and Xiaokui Xiao. 2017.
\newblock Coupled multi-layer attentions for co-extraction of aspect and
  opinion terms.
\newblock In \emph{Thirty-First AAAI Conference on Artificial Intelligence}.

\bibitem[{Wang et~al.(2016{\natexlab{b}})Wang, Huang, Zhao
  et~al.}]{wang2016attention}
Yequan Wang, Minlie Huang, Li~Zhao, et~al. 2016{\natexlab{b}}.
\newblock Attention-based lstm for aspect-level sentiment classification.
\newblock In \emph{Proceedings of the 2016 conference on empirical methods in
  natural language processing}, pages 606--615.

\bibitem[{Xu et~al.(2019)Xu, Liu, Shu, and Yu}]{xu_bert2019}
Hu~Xu, Bing Liu, Lei Shu, and Philip~S. Yu. 2019.
\newblock Bert post-training for review reading comprehension and aspect-based
  sentiment analysis.
\newblock In \emph{Proceedings of the 2019 Conference of the North American
  Chapter of the Association for Computational Linguistics}.

\end{thebibliography}
\bibliographystyle{acl_natbib}

\end{document}